\pdfoutput=1

\documentclass[11pt]{article}

\usepackage{acl}

\usepackage{times}
\usepackage{latexsym}
\usepackage{multirow, multicol}
\usepackage{algorithmic}
\usepackage[ruled,linesnumbered]{algorithm2e}
\usepackage{graphicx}
\usepackage{amsmath}
\usepackage{ulem}

\usepackage[T1]{fontenc}

\usepackage[utf8]{inputenc}

\usepackage{microtype}

\usepackage{inconsolata}

\title{Self-training Strategies for Sentiment Analysis: An Empirical Study}

\author{Haochen Liu \\
  Fidelity Investments \\
  \texttt{haochen.liu@fmr.com} \\
  \And
  Sai Krishna Rallabandi \\
  Fidelity Investments \\
  \texttt{saiKrishna.rallabandi@fmr.com} \\
  \AND
  Yijing Wu \\
  Fidelity Investments \\
  \texttt{yijing.wu@fmr.com} \\
  \And
  Parag Pravin Dakle \\
  Fidelity Investments \\
  \texttt{paragpravin.dakle@fmr.com} \\
  \And
  Preethi Raghavan \\
  Fidelity Investments \\
  \texttt{preethi.raghavan@fmr.com} \\
  }

\begin{document}
\maketitle
\begin{abstract}
Sentiment analysis is a crucial task in natural language processing that involves identifying and extracting subjective sentiment from text. Self-training has recently emerged as an economical and efficient technique for developing sentiment analysis models by leveraging a small amount of labeled data and a large amount of unlabeled data. However, given a set of training data, how to utilize them to conduct self-training makes a significant difference in the final performance of the model. We refer to this methodology as the self-training strategy. In this paper, we present an empirical study of various self-training strategies for sentiment analysis. First, we investigate the influence of the self-training strategy and hyper-parameters on the performance of traditional small language models (SLMs) in various few-shot settings. Second, we also explore the feasibility of leveraging large language models (LLMs) to help self-training. We propose and empirically compare several self-training strategies with the intervention of LLMs. Extensive experiments are conducted on three real-world sentiment analysis datasets.
\end{abstract}

\section{Introduction}
Sentiment analysis is an important and popular technique used in natural language processing (NLP) to analyze text data and determine the sentiment expressed \cite{medhat2014sentiment,chaturvedi2018distinguishing}. From social media monitoring and customer support management to customer feedback analysis, sentiment analysis has been widely applied in various daily business scenarios \cite{kumar2019fusion,bose2020sentiment}. Machine learning based sentiment detection models are usually developed via supervised learning, whose success relies on extensive, high-quality human-annotated data. However, human-labeled data is typically limited and expensive to obtain. Plus, human annotations can be noisy and require statistical filtering before usage \citep{Wang2023}. To this end, self-training is proposed to leverage a small amount of labeled data and a large amount of unlabeled data to enhance the model's performance while reducing the annotation costs \cite{kesgin2022investigating}. Self-training starts with some initial seed sentiment patterns and then uses iterative training to enlarge these patterns. It has been proven to train promising sentiment models with limited labeled data \cite{gao2014semi,van2016predicting}.

The choice of self-training strategies determines the training effect of the sentiment analysis models to a great extent. Nevertheless, they have not been studied thoroughly. In this paper, we present an empirical study on self-training strategies. 
Self-training sentiment analysis with SLMs follows an iterative two-step procedure. First, the model is initialized via supervised training on the labeled data. Second, the model makes inferences on the unlabeled data, selects the reliable instances with inferred labels, and adds them to the labeled training set. Then the model is retrained on the new labeled set, and we repeat the procedure until certain requirements are met (e.g., no more labeled data can be added). In this procedure, how to select reliable instances to add makes a big difference. Various instance selection strategies can be adopted. For example, we can decide based on the model's confidence in its prediction (e.g. the confidence score, or the entropy of the predicted probability distribution). For different tasks or datasets, the best instance selection strategy varies. In this work, we present an empirical study on the instance selection strategies of self-training for SLMs on three public sentiment analysis datasets and analyze how the choice of strategy and hyper-parameters affect the self-training performance in different few-shot settings.

With the advent of LLMs, they are extensively adopted and show promising performances in various NLP tasks, including sentiment analysis \cite{zhang2023sentiment}. They can be involved in self-training to facilitate this procedure in two modes: \textbf{subject mode} and \textbf{object mode}. In subject mode, the LLM is treated as the sentiment classifier, and the labeled or unlabeled are fed into it via prompts to improve its performance on the specific task. In the object mode, the LLM serves as an assistant to help train an SLM as the sentiment classifier. For example, the LLM can provide pseudo labels for the unlabeled data so that the SLM gets more labeled data for training. Which mode works better under different conditions? What strategies should we use? To answer these questions, we conduct experiments on three real-world sentiment datasets with two popular LLMs: Flan-UL2 and GPT-4, and summarize the empirical conclusions.

We summarize our contributions as follows: (i) we propose several instance selection strategies for self-training sentiment analysis with SLMs; (ii) we conduct an extensive comparison among various instance selection strategies for SLMs and summarize our findings on how instance selection strategies and hyper-parameters affect the efficacy of self-training for SLMs; (iii) we propose and categorize several self-training strategies for sentiment analysis models with the intervention of LLMs; (iv) we empirically compare the self-training strategies for LLMs and conclude on their applicability under different conditions.

\section{Related Works}
Sentiment analysis approaches commonly applied by the industry have experienced a transition from lexicon-based methods to machine learning based methods \cite{birjali2021comprehensive}. The latter leverages machine learning algorithms and training data to develop sentiment classification models \cite{sankar2017investigating}. In this category, various feature extraction techniques including bag of words (BoW) and distributed representations, as known as word embeddings, can be adopted. With the prosperity of deep learning and language models, the latter gradually dominates. Diverse word embedding models are proposed \cite{mikolov2013efficient,pennington2014glove,bojanowski2017enriching}, and endeavors are also conducted to improve the quality of word embeddings through statistical perspective \citep{wang2023deviance}.

The family of machine learning based sentiment analysis methods can be further divided into supervised learning, unsupervised learning, semi-supervised learning, and reinforcement learning. Supervised learning methods require high-quality labeled data for training \cite{oneto2016statistical}. In contrast, unsupervised learning models can be built using a large amount of unlabeled data, and they can handle the case that the specific sentimental classes are not given \cite{li2017learning}. Moreover, semi-supervised learning methods train the model with a few labeled data and enhance it with a large set of unlabeled data \cite{hussain2018semi,kesgin2022investigating}. Reinforcement learning methods strengthen the capability of a sentiment classifier with the trial and error mechanism \cite{rong2014auto}.

The self-training approach is one kind of semi-supervised learning method. \citet{gao2014semi} develop a self-training method where they employ multiple feature subspace-based classifiers to select useful features for sentiment classification and choose informative unlabeled samples for labeling. To alleviate the issue of errors being self-reinforcing in self-training, \citet{hong2014competitive} propose to create three models based on the models' outputs and choose the best one. \citet{hajmohammadi2015combination} introduce a novel framework that combines self-training with active learning for cross-lingual sentiment classification. In addition, \citet{van2016predicting} explore when self-training can improve the performance of sentiment analysis models. They find that the similarity among the labeled, unlabeled, and evaluation data can determine whether self-training is beneficial.
\section{Self-training with SLMs}
We first investigate self-training sentiment analysis with SLMs. In this section, we introduce the base SLM used for sentiment analysis, the general self-training procedure, and the instance selection strategies we explored.

\subsection{The Base Model}
We employ the pre-trained robustly optimized BERT approach (RoBERTa) \cite{liu2019roberta} as the base sentiment classifier. RoBERTa is a powerful model that shares the same architecture as BERT \cite{devlin2018bert}, with adjustments made upon the latter, including removing BERT's next-sentence objective and being trained with a larger batch size and learning rate. The RoBERTa model has been widely used in text classification tasks and achieves promising performances.

\subsection{General Self-training Procedure}
This study considers the sentiment classification task with three labels: positive, negative, and neutral. Given a labeled training set $\mathcal{T}=\{(s_i, c_i)\}_{i=1}^{N}$ and an unlabeled training set $\mathcal{T'}=\{s_i\}_{i=1}^{N'}$, where $N<<N'$, the task is to train a sentiment classifier $\mathbf{M}$ under an instance selection strategy $S$, and an iteration termination condition $R$.

\begin{algorithm}[h]\small

\KwIn{Labeled training set $\mathcal{T}=\{(s_i, c_i)\}_{i=1}^{N}$, unlabeled training set $\mathcal{T'}=\{s_i\}_{i=1}^{N'}$, an instance selection strategy $S$, an iteration termination condition $R$.}
\KwOut{a sentiment classifier $\mathbf{M}$.}

Initialize the sentiment classifier $\mathbf{M}$ by training it on the labeled training set $\mathcal{T}$.

\Repeat{The iteration termination condition $R$ is satisfied}{
    For each instance $s_i \in \mathcal{T'}$, use the current model $\mathbf{M}$ to infer a pseudo-label $c'_i$\\
    Select the instances $\mathcal{T^*}=\{(s_i, c'_i)|S\textrm{ is satisfied}\}$ according the instance selection strategy $S$\\
    Add the instances to the labeled set $\mathcal{T} = \mathcal{T} \cup \mathcal{T^*}$\\
    Remove the instances from the unlabeled set $\mathcal{T'} = \mathcal{T'} \backslash \mathcal{T^*}$\\
    Retrain the model $\mathbf{M}$ on the current labeled set $\mathcal{T}$
}

\caption{{\bf Self-training procedure} \label{alg:selftrain}}

\end{algorithm}

The general procedure of self-training in sentiment analysis is presented in Algorithm \ref{alg:selftrain}. First, we train the sentiment classifier on the labeled training set $\mathcal{T}$ via supervised learning (line 1). Then we update the model iteratively (lines 2-8) by repeating two steps: incorporating more labeled data from unlabeled data (lines 3-6) and retraining the model with the updated labeled set (line 7). Specifically, we carry out inference on all the instances in the unlabeled set with the current model (line 3); select the reliable instances that satisfy the given instance selection strategy (line 4), and add them into the labeled set (line 5), meanwhile, remove them from the unlabeled set (line 6). The training loop stops when a certain termination condition is satisfied, e.g., no more unlabeled instances can be added, or the model's performance doesn't improve for a certain number of consecutive epochs (line 8).
\subsection{Instance Selection Strategies}
In this section, we propose several heuristic instance selection strategies. The instance selection strategies determine which instances in the unlabeled data can be used for training with the inferred pseudo-labels. The principle of selecting such instances is to ensure the reliability of the pseudo-labels -- correct labels will enhance the reasoning capability of the model and improve its generalization ability. In contrast, wrong labels bring negative impacts on the model.

\subsubsection{Threshold-based}
The threshold-based methods judge whether an instance with inferred pseudo-labels is good to use by comparing its reliability measurement with a pre-defined threshold $t$.

\textbf{Confidence Score}: the strategy selects instances whose pseudo-label's predicted probability (i.e. confidence score) is above the given threshold $t$. A high predicted probability implies the model is confident with its prediction, which means the inferred label is expected to be accurate.

\textbf{Distribution Entropy}: the strategy selects instances whose predicted probability distribution's entropy is lower than the given threshold $t$. A low-entropy probability distribution implies a more certain prediction, which means the inferred label is more reliable.

\subsubsection{Max/Min-based}
The max/min-based methods consider the same two measurements as the threshold-based methods. However, the max/min-based methods select the instances with top-$k$ reliability measurement scores in the unlabeled set and add them into the labeled set with their inferred labels.

\textbf{Confidence Score}: select $k$ instances with maximal confidence scores in the unlabeled set. 

\textbf{Distribution Entropy}: select $k$ instances with minimal distribution thresholds in the unlabeled set. 

\subsubsection{Soft Label}
Unlike the above two methods, where a pseudo-label is explicitly inferred and added to the labeled data, the soft-label method uses the inferred probability distribution of the unlabeled instances as the signals for training the model. For an unlabeled instance $s_i \in \mathcal{T'}$, we treat the inferred distribution $\hat{p}$ as the target, and train the model by optimizing the Kullback–Leibler (KL) divergence between the predicted distribution $p$ and the target distribution $\hat{p}$: $L=KL(p, \hat{p})$.

\begin{table*}[t]
    \small
    \caption{Empirical comparison among different instance selection strategies on the \textbf{LDC} and the \textbf{MOSEI} datasets in various n-shot settings. The average F1 scores of 3 runs are reported. As a reference, the model trained on all available labeled data can achieve F1 scores of $0.803$ and $0.522$ on the LDC and the MOSEI datasets, respectively.}
	\label{tab:exp1}
	\begin{tabular}{l|c|c|c|c|c|c||c|c|c|c|c|c}
        \hline
        \hline
         & \multicolumn{6}{c||}{\textbf{LDC}} & \multicolumn{6}{c}{\textbf{MOSEI}} \\
        \hline
		\textbf{n-shot} & \textbf{5} & \textbf{10} & \textbf{15} & \textbf{20} & \textbf{25} & \textbf{30} & \textbf{5} & \textbf{10} & \textbf{15} & \textbf{20} & \textbf{25} & \textbf{30} \\
		\hline
		SL & 0.234 & 0.298 & 0.296 & 0.557 & 0.601 & 0.650 & 0.259 & 0.324 & 0.400 & 0.416 & 0.436 & 0.458 \\
         RS & 0.257 & 0.189 & 0.292 & 0.547 & 0.612 & 0.670 & 0.276 & 0.284 & 0.252 & 0.386 & 0.470 & 0.448 \\
		\hline
		 Conf. Thr. & 0.338 & 0.263 & 0.368 & 0.613 & 0.649 & 0.722 & 0.275 & 0.324 & 0.408 & 0.425 & 0.470 & 0.471 \\
          Ent. Thr. & 0.338 & 0.263 & 0.368 & 0.625 & 0.651 & 0.710 & 0.259 & 0.324 & 0.400 & 0.416 & 0.457 & 0.475 \\
          Max Conf. & 0.193 & 0.198 & 0.104 & 0.562 & 0.629 & 0.661 & 0.100 & 0.214 & 0.324 & 0.221 & 0.417 & 0.366 \\
          Min Ent. & 0.194 & 0.190 & 0.118 & 0.525 & 0.596 & 0.581 & 0.098 & 0.219 & 0.349 & 0.275 & 0.424 & 0.351  \\
          Soft Labels & 0.453 & 0.472 & 0.502 & 0.546 & 0.627 & 0.667 & 0.321 & 0.319 & 0.430 & 0.321 & 0.445 & 0.450 \\
         \hline
         \hline
	\end{tabular}
\end{table*}

\begin{table*}[t]
    \small
    \caption{Empirical comparison among different instance selection strategies on the \textbf{Financial Phrasebank} dataset in various n-shot settings. The average F1 scores of 3 runs are reported. As a reference, the model trained on all available labeled data can achieve F1 scores of $0.972$ and $0.878$ on all agree and 50 agree datasets, respectively.}
	\label{tab:exp2}
	\begin{tabular}{l|c|c|c|c|c|c||c|c|c|c|c|c}
        \hline
        \hline
         & \multicolumn{6}{c||}{\textbf{Financial Phrasebank (All Agree)}} & \multicolumn{6}{c}{\textbf{Financial Phrasebank (50 Agree)}} \\
        \hline
		\textbf{n-shot} & \textbf{5} & \textbf{10} & \textbf{15} & \textbf{20} & \textbf{25} & \textbf{30} & \textbf{5} & \textbf{10} & \textbf{15} & \textbf{20} & \textbf{25} & \textbf{30} \\
		\hline
		SL & 0.712 & 0.739 & 0.762 & 0.824 & 0.876 & 0.908 & 0.204 & 0.513 & 0.510 & 0.631 & 0.675 & 0.710 \\
         RS & 0.679 & 0.753 & 0.790 & 0.823 & 0.866 & 0.887 & 0.122 & 0.495 & 0.543 & 0.632 & 0.681 & 0.741 \\
		\hline
		 Conf. Thr. & 0.680 & 0.824 & 0.780 & 0.815 & 0.833 & 0.910 & 0.235 & 0.375 & 0.568 & 0.612 & 0.708 & 0.727 \\
          Ent. Thr. & 0.712 & 0.776 & 0.782 & 0.866 & 0.867 & 0.899 & 0.235 & 0.451 & 0.497 & 0.644 & 0.701 & 0.717 \\
          Max Conf. & 0.632 & 0.730 & 0.755 & 0.868 & 0.866 & 0.904 & 0.121 & 0.482 & 0.133 & 0.635 & 0.676 & 0.663 \\
          Min Ent. & 0.647 & 0.731 & 0.760 & 0.847 & 0.866 & 0.900 & 0.108 & 0.369 & 0.219 & 0.603 & 0.695 & 0.708 \\
          Soft Labels & 0.686 & 0.721 & 0.719 & 0.855 & 0.869 & 0.935 & 0.282 & 0.571 & 0.593 & 0.605 & 0.716 & 0.700 \\
         \hline
         \hline
	\end{tabular}
\end{table*}

\section{Experiments I: SLMs}
This section presents our experiments of various instance selection strategies for SLMs on three public datasets: (i) the multimodal corpus for sentiment analysis released by the Linguistic Data Consortium (LDC) \cite{chen2020large}; (ii) the CMU multimodal opinion sentiment and emotion intensity (MOSEI) dataset \cite{zadeh2018multi}; and (iii) the Financial Phrasebank dataset (FP) \cite{Malo2014GoodDO}.
These three datasets involve sentiment classification tasks with different granularities: the LDC and FP datasets contain shorter, sentence-level texts while the MOSEI dataset consists of longer, paragraph-level texts.

Through the experiments, we seek to investigate the following research questions: (i) How does each instance strategy selection perform for SLMs under different settings? (ii) How does each hyper-parameter impact the performance of the self-training procedure?

\subsection{Datasets}
In this section, we introduce the details of the public datasets used in our experiments.

\subsubsection{The LDC Dataset}
The LDC dataset is extended from the Switchboard-1 telephone speech corpus. It contains the transcripts of 49,500 speech segments of 140 hours of audio. Each segment is a sentence, and was labeled by 3 human annotators into three sentiment categories: positive, neutral, and negative.

\subsubsection{The MOSEI Dataset}
The MOSEI dataset is a multimodal opinion sentiment analysis dataset, which consists of monologue videos from 1,000 YouTube speakers. In total, 3,293 videos are transcribed to texts that contain multiple sentences. Like the LDC dataset, each text was labeled by human annotators into three sentiment categories: positive, neutral, and negative.

\subsubsection{The Financial Phrasebank Dataset}
The Financial PhraseBank dataset is a widely used dataset for financial NLP tasks, particularly financial sentiment analysis. It contains over 10,000 sentences collected from financial news articles, annotated by finance professionals with respect to their sentiment polarity (positive, negative, or neutral). The dataset covers a diverse range of financial topics, such as corporate strategy, financial performance, and market trends. We use two splits of this dataset for experiments: (i) all agree: this split contains sentences for which all annotators achieve an agreement regarding the sentiment polarity. It is ideal for evaluating the performance of models in scenarios where the sentiment is relatively clear and unambiguous; (ii) 50 agree: this split contains sentences for which more than 50\% annotations achieve an agreement. Evaluating models on this split can help assess their ability to handle ambiguous or conflicting sentiment cues.

\subsubsection{Data Distributions}
The category distributions of the datasets we use are as follows.

\begin{itemize}
    \item LDC: 5658 positive, 2578 negative, 10106 neutral
    \item MOSEI: 1509 positive, 432 negative, 693 neutral
    \item FP Allagree: 514 positive, 266 negative, 1257 neutral
    \item FP 50agree: 1239 positive, 533 negative, 2589 neutral
\end{itemize}

\subsection{Experimental Settings}
We conduct experiments on various $n$-shot settings, where $n$ indicates the number of labeled instances of each class given in the labeled training set $\mathcal{T}$. Specifically, we report the results under 5, 10, 15, 20, 25, and 30-shot settings.

We compare the instance selection strategies of interest with two baseline methods: supervised learning (SL) and random sampling (RS). The former uses only the $n$-shot labeled instances for supervised learning. The latter adopts a random strategy for selecting instances in self-training: a batch of unlabeled instances is randomly picked at each iteration.

\subsection{Implementation Details}
\label{sec:imp}
We use the pre-trained Roberta-base model \cite{2019RoBERTa} as our base classifier. It has 125M trainable parameters. We do the experiments on NVIDIA Tesla K80 GPUs. Each self-training experiment takes no more than 10 minutes. The initial learning rate is set as $8e-6$. The model initializing and retraining steps stop when the model's performance on the validation set doesn't improve for $2$ consecutive epochs. After that, a batch of at most $1,000$ unlabeled instances selected by the strategies are added into the labeled set (if there are less than $1,000$ instances that can be selected, then we select as many as possible.) The self-training terminates when no more unlabeled data can be selected.

For both the LDC and the MOSEI datasets, $20\%$ data are randomly picked as the test set, and the remaining $80\%$ data are used for training. Within the training data, $n$ instances are sampled as the labeled data for model initialization under the $n$-shot setting; while the rest of the training data are used as unlabeled data for self-training.

\subsection{Performance Comparison}
We summarize the experimental results of various instance strategies on the three datasets in Table \ref{tab:exp1} and Table \ref{tab:exp2}. We make the following observations. First, compared with the supervised learning baseline, self-training can enhance the model's performance by utilizing unlabeled data, when enough labeled data are provided ($n\geq 20$) at the beginning for model initialization. Second, self-training can not always help when there are fewer labeled data, because the performance of the initialized model determines the quality of the new instances added from the unlabeled set to the training set in the following self-training steps. Third, different instance selection strategies have varying performances. In most cases (except for the FP All Agree dataset, where the instances are less ambiguous), the soft label method performs the best when fewer labeled data are given. The soft label method doesn't explicitly predict a pseudo-label for self-training but uses the predicted probability distribution as the supervised signal. It has a greater fault tolerance by avoiding errors caused by mispredicted pseudo-labels when the model is not well initialized with limited labeled data. On the contrary, the confidence/entropy threshold strategies work better when more labeled data are given. It is because when the model is well initialized, the threshold-based strategies can help us find instances with reliable pseudo-labels, so as to improve the model with accurate additional training data in self-training.

\begin{figure}[t]
    \centering
    \includegraphics[width=230pt]{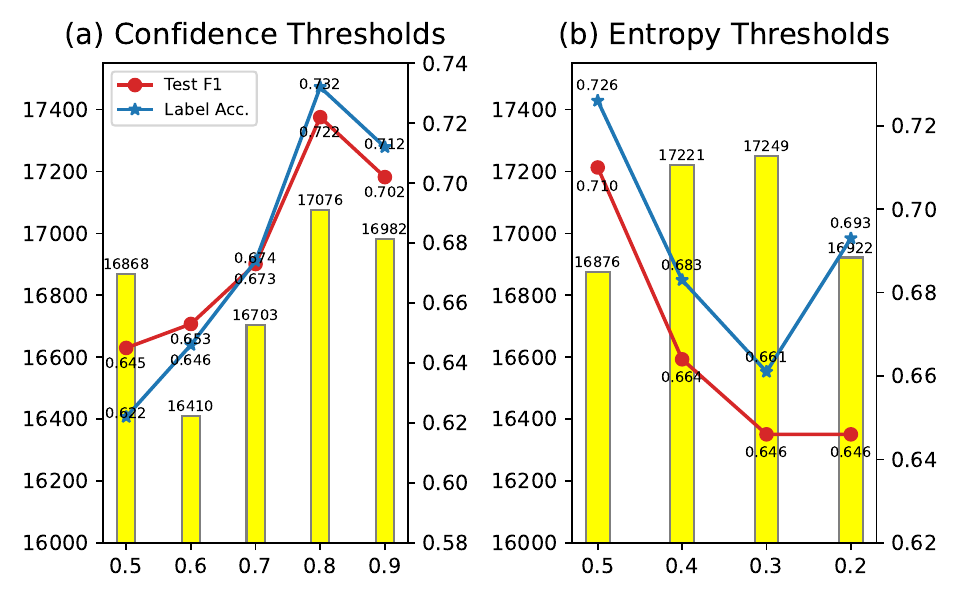}
    \caption{The x-axis indicates the threshold; the yellow bars represent the final number of unlabeled instances added to the training set; the blue line indicates the accuracy of inferring unlabeled instances; the red line indicates the F1 score of the well-trained model on the test set.} 
    \label{fig:param}
\end{figure}

\begin{figure}[t]
    \centering
    \includegraphics[scale=0.57]{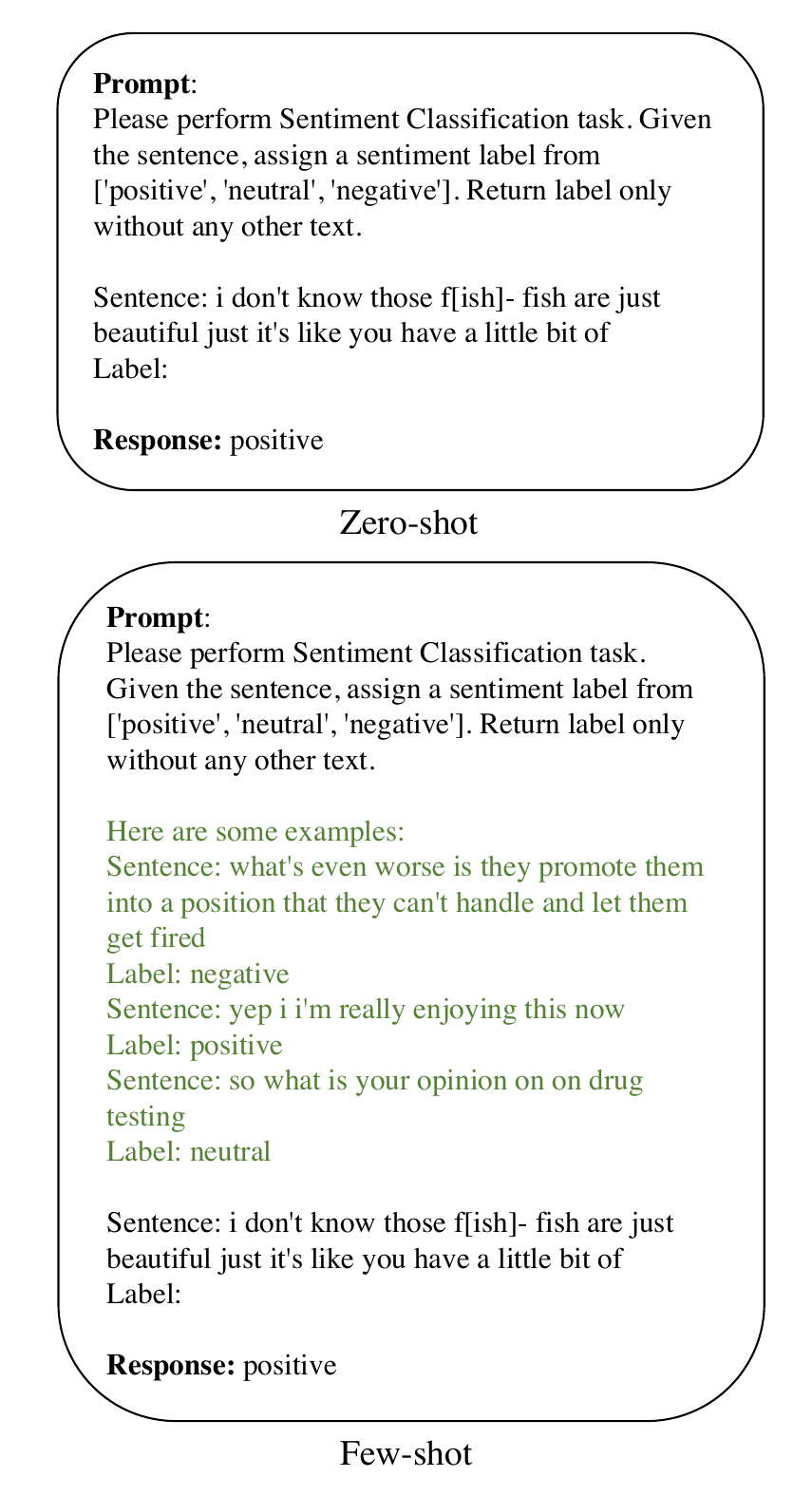}
    \caption{The prompts used for querying LLMs in the zero-shot and few-shot settings.}
    \label{fig:prompt}
\end{figure}

\subsection{Hyper-parameter Analysis}
The threshold-based methods perform the best when a considerable amount of labeled data is given. We further investigate how the choice of thresholds impacts the performance of the confidence- and entropy-threshold methods, on the LDC data. In figure \ref{fig:param}, the experimental results under the 20-shot setting are reported. First, we find that along with the change of the thresholds, the number of unlabeled instances added to the training set doesn't show a monotonous trend as expected, i.e. a stricter threshold leads to fewer data to add. In fact, sometimes a strict threshold can select unlabeled data of higher quality in the early stage of self-training, then a more accurate model is obtained, so that more unlabeled instances can be inferred with high confidence and selected in subsequent iterations. Second, we find that in the self-training process, the accuracy of the inferred pseudo-labels shows a strong correlation with the model's final performance, while the amount of selected unlabeled instances is not important. It suggests we focus more on ensuring the quality of newly added data during self-training, instead of the quantity.


\section{Self-training with LLMs}
LLMs are trained on extremely huge corpora, which endow them with promising capability in many tasks and domains for which they have not been specifically trained. We can leverage LLMs to facilitate a certain sentiment analysis task under the self-training setting (i.e. a small set of labeled data and a large set of unlabeled data are given) in two modes: subject mode and object mode.

\begin{table*}[t]
    \centering
    \small
    \caption{The performances of the \textbf{Sub} strategy. In the 5-shot setting, we try three different sets of examples to provide to the LLM and report the average result with a 95\% confidence interval. ``NA'' indicates the unavailable results due to the input limitation of LLMs.}
	\label{tab:sub}
	\begin{tabular}{l|c|c||c|c||c|c||c|c}
        \hline
        \hline
         & \multicolumn{4}{c||}{\textbf{Flan-UL2}} & \multicolumn{4}{c}{\textbf{GPT-4}} \\
        \hline
        \hline
         & \multicolumn{2}{c||}{\textbf{LDC}} & \multicolumn{2}{c||}{\textbf{MOSEI}} & \multicolumn{2}{c||}{\textbf{LDC}} & \multicolumn{2}{c}{\textbf{MOSEI}} \\
        \hline
		\textbf{n-shot} & \textbf{0} & \textbf{5} & \textbf{0} & \textbf{5} & \textbf{0} & \textbf{5} & \textbf{0} & \textbf{5} \\ \hline
         \textbf{Accuracy} & 0.635 & 0.680$\pm$0.013 & 0.542 & 0.334$\pm$0.531 & 0.731 & 0.690$\pm$0.034 & 0.546 & NA \\
         \textbf{F1} & 0.630 & 0.685$\pm$0.015 & 0.509 & 0.191$\pm$0.495 & 0.729 & 0.692$\pm$0.033 & 0.554 & NA \\
		 \hline
         \hline
         & \multicolumn{2}{c||}{\textbf{FP (All Agree)}} & \multicolumn{2}{c}{\textbf{FP (50 Agree)}} & \multicolumn{2}{c||}{\textbf{FP (All Agree)}} & \multicolumn{2}{c}{\textbf{FP (50 Agree)}} \\
        \hline
		\textbf{n-shot} & \textbf{0} & \textbf{5} & \textbf{0} & \textbf{5} & \textbf{0} & \textbf{5} & \textbf{0} & \textbf{5} \\ \hline
         \textbf{Accuracy} & 0.912 & 0.959$\pm$0.006 & 0.804 & 0.852$\pm$0.021 & 0.899 & 0.943$\pm$0.048 & 0.759 & 0.781$\pm$0.090 \\
         \textbf{F1} & 0.913 & 0.959$\pm$0.007 & 0.806 & 0.852$\pm$0.020 & 0.900 & 0.943$\pm$0.048 & 0.765 & 0.784$\pm$0.085 \\
		 \hline
         \hline
	\end{tabular}
\end{table*}

\begin{table*}[t]
    \small
    \caption{The performances of the \textbf{Obj} strategy in zero-shot and 5-shot settings. The ``Label.'' columns show the accuracy of the LLM inferring unlabeled instances. The ``Infer.'' columns show the F1 score of the SLM trained on pseudo-labels inferring the test instances.}
	\label{tab:obj}
	\begin{tabular}{l||c|c||c|c||c|c||c|c}
        \hline
        \hline
         & \multicolumn{4}{|c||}{\textbf{Flan-UL2}} & \multicolumn{4}{c}{\textbf{GPT-4}} \\
        \hline
        \hline
         & \multicolumn{2}{c||}{\textbf{0-shot}} & \multicolumn{2}{c||}{\textbf{5-shot}} & \multicolumn{2}{c||}{\textbf{0-shot}} & \multicolumn{2}{c}{\textbf{5-shot}} \\ \hline
         & Label. & Infer. & Label. & Infer. & Label. & Infer. & Label. & Infer. \\ \hline
        \textbf{LDC} & 0.626 & 0.154 & 0.678$\pm$0.014 & 0.703$\pm$0.030 & 0.710 & 0.712 & 0.685$\pm$0.006 &  0.706$\pm$0.043 \\ \hline
        \textbf{MOSEI} & 0.542 & 0.417 & 0.333$\pm$0.532 & 0.191$\pm$0.495 & 0.478 & 0.474 & NA & NA\\ \hline
        \textbf{FP (All Agree)} & 0.913 & 0.910 &  0.950$\pm$0.007 & 0.935$\pm$0.029 & 0.902 & 0.920 & 0.925$\pm$0.035 & 0.928$\pm$0.015 \\ \hline
        \textbf{FP (50 Agree)} & 0.781 & 0.795 & 0.825$\pm$0.009 & 0.836$\pm$0.045 & 0.758 & 0.775 & 0.770$\pm$0.039 & 0.802$\pm$0.041\\
         \hline
         \hline
	\end{tabular}
\end{table*}

\begin{table*}[t]
    \centering
    \small
    \caption{The performances of the \textbf{Obj-Conf} strategy. ``\# Train'' indicates the number of unlabeled instances whose pseudo-labels the LLM is confident with, out of the total number of unlabeled instances.}
	\label{tab:obj_conf}
	\begin{tabular}{l||c|c|c||c|c|c}
        \hline
        \hline
         & \multicolumn{6}{c}{\textbf{Flan-UL2}} \\
        \hline
        \hline
         & \multicolumn{3}{c||}{\textbf{0-shot}} & \multicolumn{3}{c}{\textbf{5-shot}} \\ \hline
         & \# Train & Label. & Infer. & \# Train & Label. & Infer. \\ \hline
        \textbf{LDC} & 325/18342 & 0.074 & 0.381 & 407.3/18327 & 0.328$\pm$0.260 & 0.402$\pm$0.056 \\ \hline
        \textbf{MOSEI} & 180/2634 & 0.200 & 0.109 & NA & NA & NA \\ \hline
        \textbf{FP (All Agree)} & 63/2037 & 0.841 & 0.368 & 81.3/2022 & 0.987$\pm$0.008 & 0.098$\pm$0.000 \\ \hline
        \textbf{FP (50 Agree)} & 83/4361 & 0.747 & 0.530 & 119.0/4346 & 0.936$\pm$0.012 & 0.159$\pm$0.145 \\
         \hline
         \hline
         & \multicolumn{6}{c}{\textbf{GPT-4}} \\
        \hline
        \hline
         & \multicolumn{3}{c||}{\textbf{0-shot}} & \multicolumn{3}{c}{\textbf{5-shot}} \\ \hline
         & \# Train & Label. & Infer. & \# Train & Label. & Infer. \\ \hline
        \textbf{LDC} & 17687/18342 & 0.711 & 0.721 & 13446.3/18327 & 0.686$\pm$0.006 & 0.709$\pm$0.013 \\ \hline
        \textbf{MOSEI} & 2620/2634 & 0.480 & 0.371 & NA & NA & NA \\ \hline
        \textbf{FP (All Agree)} & 2015/2037 & 0.894 & 0.917 & 2011.3/2022 & 0.925$\pm$0.033 & 0.915$\pm$0.017 \\ \hline
        \textbf{FP (50 Agree)} & 4328/4361 & 0.757 & 0.771 & 4330.0/4346 & 0.770$\pm$0.039 & 0.795$\pm$0.076 \\
         \hline
         \hline
	\end{tabular}
\end{table*}

\subsection{Subject Mode}
In the subject mode, we treat the LLM itself as the sentiment classifier. We can either directly ask the LLM to perform the sentiment analysis task with appropriate prompts (zero-shot setting) or provide the LLM with a few instances (few-shot setting) and true labels, and then ask it to do the inference. We refer to the subject mode as \textbf{Sub} strategy in the following experiments.

\textbf{Prompting Strategy.} To make the experiment results robust, following \citet{zhang2023sentiment}, we ask GPT-4 to generate the prompt while ensuring the prompts are as simple and clear as possible, and we use consistent prompts for different experiments. Such a prompting strategy helps us make an objective evaluation of various models. In Figure \ref{fig:prompt}, we show the prompts we used for LLM experiments in the zero-shot and the few-shot settings.

\subsection{Object Mode}
In the object mode, we ask the LLM to infer the pseudo labels of unlabeled data, and then use them as an augmentation of labeled data to train an SLM as the sentiment classifier. However, the predictions of the LLM are not always precise. Thus, we can ask the LLM to estimate the confidence in its predictions and decide whether we should incorporate the corresponding instance for training. We propose three strategies:

\begin{itemize}
    \item \textbf{Obj}: An LLM is employed to predict the labels for all unlabeled instances. The inferred labels are incorporated as the pseudo-labels for subsequent SLM training.
    \item \textbf{Obj-Conf}: An LLM is employed to predict the labels for all unlabeled instances, as well as a binary indicator presenting whether the LLM is confident with its prediction. The inferred labels that the LLM is confident with are incorporated for subsequent SLM training.
    \item \textbf{Obj-Conf-Score}: An LLM is employed to predict the labels for all unlabeled instances, as well as a confidence score of its prediction ranging from 0 to 1. The inferred labels whose confidence score is higher than a threshold are incorporated for subsequent SLM training.
\end{itemize}



\section{Experiments II: LLMs}
We conduct experiments on two popular LLMs: Flan-UL2 \cite{2022UL2} and GPT-4 \cite{openai2023gpt}

\subsection{Performance Comparison}
\textbf{Sub Strategy.} The performances of the Sub strategy are presented in Table \ref{tab:sub}. We observe that LLMs can perform well on sentiment analysis tasks even if no or few labeled data are available. Specifically, when no labeled data is given (zero-shot), GPT-4 can achieve better performances than all the instance selection strategies for SLMs in 5-30 shot settings on LDC and MOSEI datasets, which demonstrates the excellent capability of GPT-4 on unseen tasks due to the huge corpus it was trained on and its enormous model size \cite{openai2023gpt}. GPT-4 is superior to Flan-UL2 on LDC and MOSEI datasets, while the latter outperforms the former on the Finance Phrasebank dataset. What's more, interestingly, we find that a few labeled examples cannot always help LLMs. The performance of Flan-UL2 drops and becomes unstable on MOSEI when 5 examples of each sentiment class are provided. This is because text instances in this dataset are long, which leads to a verbose prompt that disturbs the model's predictions. GPT-4's performance also gets worse on the LDC dataset in the 5-shot setting. Both the results of the two LLMs on the Finance Phrasebank dataset get improved when a few examples are given. The observations above show that when an LLM is competent enough for the sentiment analysis task in an open domain (e.g. the LDC and the MOSEI datasets), providing a few examples may lead to the LLMs being biased on the examples, which undermines its generalization capability. On the contrary, in a specialized domain (e.g. the Finance Phrasebank dataset), providing examples is more likely to improve the prediction capability of LLMs in this domain.

As a reference, we add an experiment on the Finance Phrasebank dataset, where we fine-tune Flan-UL2 on the complete training set, and evaluate it on the test set. The F1 scores on the All Agree split and the 50 Agree split are 0.978 and 0.882, respectively. The results are better than those of small language models trained on the complete training set (0.972 and 0.878), which demonstrates that prior knowledge of LLMs is helpful for the sentiment classification task.

\begin{figure*}[t]
    \centering
    \includegraphics[width=450pt]{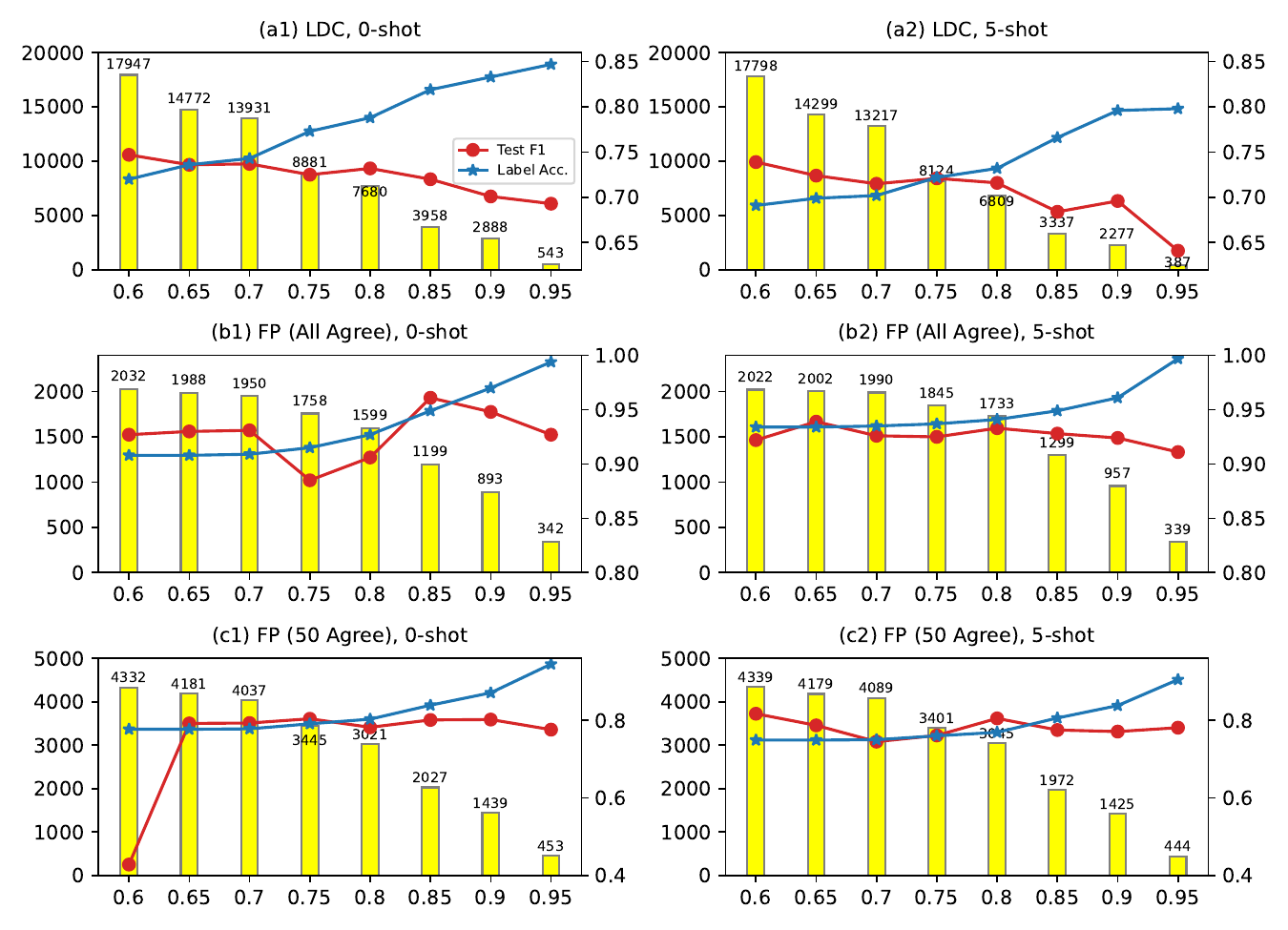}
    \caption{The performances of GPT-4 with the \textbf{Obj-Conf-Score} strategy. The x-axis indicates the thresholds of the confidence scores; the yellow bars represent the number of inferred instances selected for training; the blue line indicates the accuracy of the LLM inferring unlabeled instances; the red line indicates the F1 score of the well-trained SLM on the test set.} 
    \label{fig:obj_conf_score}
\end{figure*}

\textbf{Obj Strategy.} The results of the Obj strategy are shown in Table \ref{tab:obj}. First, the observations on the performance differences between the 0-shot and 5-shot settings are the same as the Sub strategy. Second, comparing Table \ref{tab:obj} with Table \ref{tab:sub}, we find that the performance of an SLM trained with unlabeled data with pseudo labels provided by an LLM is worse than that of the LLM itself. As we can see, the pseudo-labels inferred by LLMs are not accurate enough to train an SLM for the sentiment analysis task in a specific domain.

\textbf{Obj-Conf Strategy.} Given that the pseudo labels predicted by LLMs may not be accurate, a possible solution is to ask the LLM to estimate the confidence of its predictions and use only the confident instances for training the SLM. Table \ref{tab:obj_conf} shows the performances of the Obj-Conf strategy. We observe that Flan-UL2 is confident with only a few predictions it made, while GPT-4 is confident with most of its predictions. However, we find that the labeling accuracy of the instances the LLMs are confident with is not obviously higher than that of the Obj strategy, which means that it's hard for LLMs to provide objective and correct binary estimations of their confidence in their predictions. 

\textbf{Obj-Conf-Score Strategy.} In the Obj-Conf-Score strategy, we alternatively ask the LLM to estimate its confidence by a numeric score at a scale of 0 to 1. Flan-UL2 fails to understand the prompt to give the confidence scores as expected so we only report the results of GPT-4. Figure \ref{fig:obj_conf_score} shows how the performances of GPT-4 change along with the increase of the confidence score thresholds. First, we can see that as the confidence score thresholds increase, fewer unlabeled data with pseudo-labels are selected for training the SLM, and the labeling accuracy of selected instances rises accordingly. It demonstrates that GPT-4 is able to estimate its confidence in a quantitative form. Second, the performance of the resulting SLM fluctuates as the confidence threshold changes, and achieves the best when the threshold is 0.8-0.85. The threshold should be chosen carefully to reach a trade-off between the accuracy and the number of instances with pseudo-labels we select for training the SLM. Based on our observations in the experiments, selecting an appropriate threshold for Obj-Conf-Score is tricky since it depends on both the LLM and the dataset. Different LLMs give confidence scores in different scales; and the trade-off point between the accuracy and the number of instances varies for different datasets. Our empirical suggestion is that on the premise of keeping a certain amount of training samples (e.g. 1000), we choose the threshold that maximizes the accuracy. Third, we observe a sharp lift for the F1 score in Figure 3 c(1) when the threshold changes from 0.6 to 0.65. This is because some error cases that pass the 0.6 confidence threshold negatively affect the performances of the trained SLM. This observation shows that sometimes a few error training samples can lead to significant performance drops in the self-training setting. Finally, experiments show that when an appropriate threshold is used, the Obj-Conf-Score strategy can achieve the best performance among all the self-training strategies for sentiment analysis.

\section{Conclusion}
In this study, we present an empirical study on self-training strategies for the sentiment analysis task. We first propose several heuristic instance selection strategies for self-training with SLMs, and conduct an evaluation of them under different few-shot settings. Second, we make endeavors to leverage LLMs to help self-training. We propose and evaluate several self-training strategies with the intervention of LLMs. Based on the experiments on three public datasets, we compare different self-training strategies, discuss their applicability under various conditions, and analyze the influence of hyper-parameters on their performances. The work serves as an empirical study to assist practitioners in selecting appropriate strategies to construct sentiment analysis models when limited annotated data is available.
\section{Limitations}
The quality of the outputs of an LLM is susceptible to the prompts \cite{2021Fantastically}, which means that the empirical experiment results may vary if different prompts are used. In this study, we have tried our best to control the influence of prompts on the experiment results by using simple, precise, LLM-generated prompts, in order to reach robust and reliable conclusions. In future work, we plan to further investigate how prompt variation affects the empirical results.

\section*{Acknowledgements}
We would like to express our appreciation to Ms. Chaitra Hegde for her initial research on emotion and sentiment analysis at Fidelity Investments. Her preliminary contributions inspired this study.

\bibliography{main}

\end{document}